\documentclass[10pt,twocolumn,letterpaper]{article}

\usepackage[pagenumbers]{cvpr}
\usepackage{graphicx}
\usepackage{amsmath}
\usepackage{amssymb}
\usepackage{booktabs}
\usepackage{multirow}
\usepackage[title]{appendix}
\usepackage[group-separator={,}, group-minimum-digits=4]{siunitx}
\usepackage{tablefootnote}
\usepackage[para,symbol]{footmisc}

\usepackage[pagebackref,breaklinks,colorlinks]{hyperref}

\usepackage[capitalize]{cleveref}
\crefname{section}{Sec.}{Secs.}
\Crefname{section}{Section}{Sections}
\Crefname{table}{Table}{Tables}
\crefname{table}{Tab.}{Tabs.}

\begin{document}
\title{
LAVT: Language-Aware Vision Transformer for \\ Referring Image Segmentation
}
\author{Zhao Yang\textsuperscript{1}\hspace{-0.014cm}\footnotemark[1]~,
        Jiaqi Wang\textsuperscript{2}\hspace{0.01cm}\footnotemark[1]~,
        Yansong Tang\textsuperscript{5,1}\footnotemark[2]~,
        Kai Chen\textsuperscript{2,4},
	    Hengshuang Zhao\textsuperscript{3,1},
	    Philip H.S. Torr\textsuperscript{1}\\
\textsuperscript{1}University of Oxford,
\textsuperscript{2}Shanghai AI Laboratory,
\textsuperscript{3}The University of Hong Kong,\\
\textsuperscript{4}SenseTime Research,
\textsuperscript{5}Tsinghua-Berkeley Shenzhen Institute, Tsinghua University
}

\maketitle
\footnotetext[1]{\hspace{-0.14cm}Equal contribution.\hspace{-0.1cm}}
\footnotetext[2]{\hspace{-0.14cm}Corresponding author.}

\begin{abstract}
\label{sec:abs}
Referring image segmentation is a fundamental vision-language task that aims to segment out an object referred to by a natural language expression from an image. One of the key challenges behind this task is leveraging the referring expression for highlighting relevant positions in the image. A paradigm for tackling this problem is to leverage a powerful vision-language (``cross-modal'') decoder to fuse features independently extracted from a vision encoder and a language encoder. Recent methods have made remarkable advancements in this paradigm by exploiting Transformers as cross-modal decoders, concurrent to the Transformer's overwhelming success in many other vision-language tasks. Adopting a different approach in this work, we show that significantly better cross-modal alignments can be achieved through the early fusion of linguistic and visual features in intermediate layers of a vision Transformer encoder network. By conducting cross-modal feature fusion in the visual feature encoding stage, we can leverage the well-proven correlation modeling power of a Transformer encoder for excavating helpful multi-modal context. This way, accurate segmentation results are readily harvested with a light-weight mask predictor. Without bells and whistles, our method surpasses the previous state-of-the-art methods on RefCOCO, RefCOCO+, and G-Ref by large margins.
\end{abstract}

\begin{figure}[t]
\centering
\includegraphics[width=0.99\columnwidth]{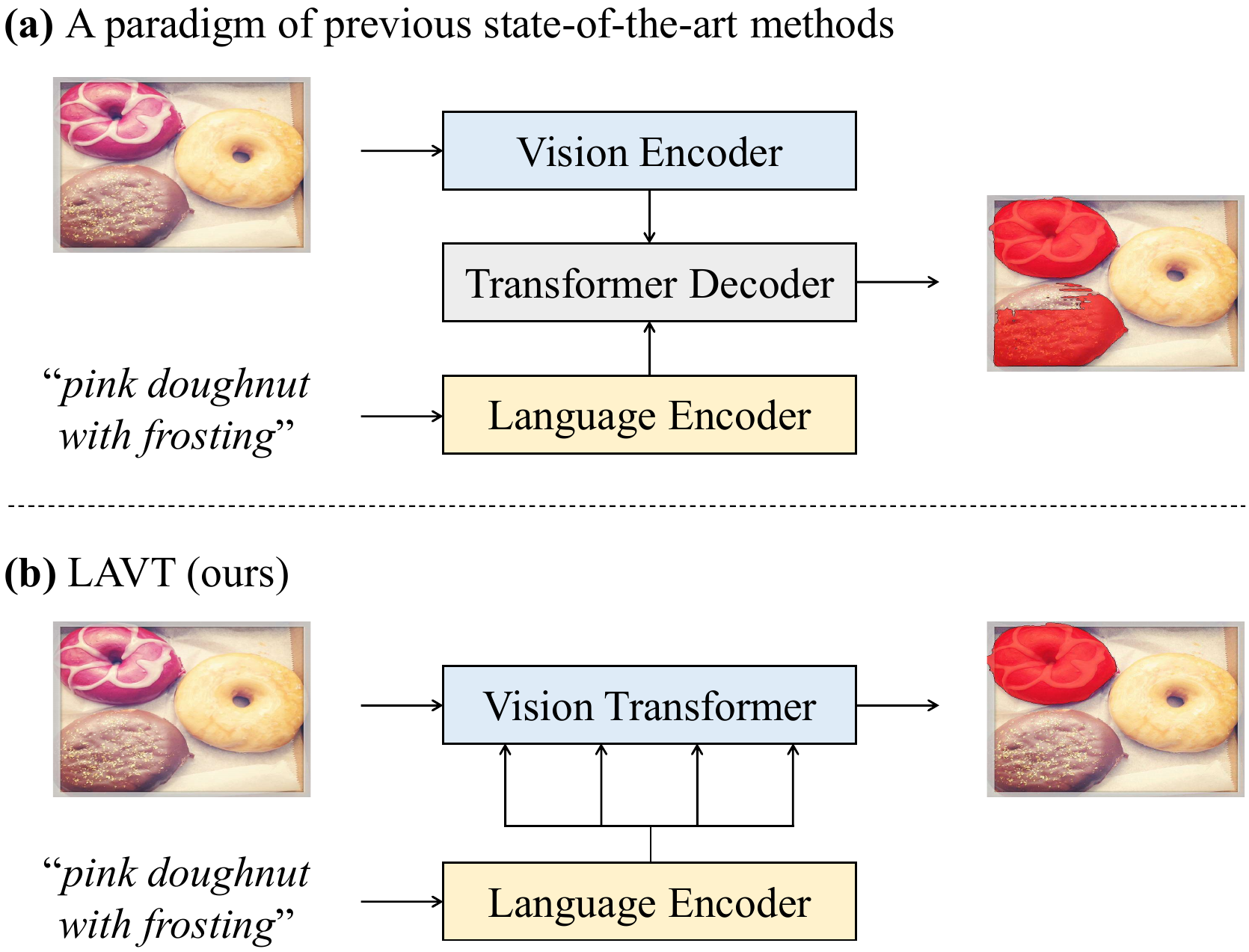}
\vspace{5pt}
\caption{The task of referring image segmentation takes one image and one text description as inputs, and predicts a mask delineating the object specified in the description. (a) The previous state-of-the-art method (\ie, VLT~\cite{Ding_2021_vlt}) leverages a vision-language Transformer decoder for cross-modal feature fusion. (b) Conversely, we propose to directly integrate linguistic information into visual features at intermediate levels of a vision Transformer network, where beneficial vision-language cues are jointly exploited. A light-weight mask predictor can thus readily replace the complicated cross-modal decoder in previous counterparts.}
\label{fig:1}
\end{figure}

\section{Introduction}
\label{sec:intro}
Given an image and a text description of the target object, referring image segmentation aims at predicting a pixel-wise mask that delineates that object~\cite{cheng2014imagespirit, hu2016segmentation}.
It yields great value for various applications such as language-based human-robot interaction~\cite{wang2019reinforced} and image editing~\cite{chen2018language}.
In contrast to conventional single-modality visual segmentation tasks based on fixed category conditions~\cite{lin2014mscoco,ade}, referring image segmentation has to deal with the much richer vocabularies and syntactic varieties of human natural languages.
In this task, the target object is inferred from a free-form expression, which includes words and phrases presenting the concepts of entities, actions, attributes, positions, \etc, organized by syntactic rules.
Therefore, the key challenge of this task is to exploit visual features that are relevant to the given text conditions.

There have been growing efforts devoted to referring image segmentation over the past few years.
A widely adopted paradigm is to first independently extract vision and language features from different encoder networks, and then fuse them together to make predictions with a cross-modal decoder.
Concretely, the fusion strategies include recurrent interaction~\cite{liu2017recurrentInter,Li2018rrn}, cross-modal attention~\cite{shi2018key,chen2019see,hu2020brinet}, multi-modal graph reasoning~\cite{huang2020CMPC}, linguistic structure-guided context modeling~\cite{hui2020linguistic}, \etc.
Recent advances (\eg,~\cite{Ding_2021_vlt}) bring performance improvements via employing a cross-modal Transformer~\cite{attention-all-you-need} decoder (illustrated in~\cref{fig:1} (a)) to learn more effective cross-modal alignments, which is in concurrence with Transformer’s overwhelming success in many other vision-language tasks~\cite{clip, hu2021unit, ViLBERT, clipbert}.

Although great progress has been achieved, the potentiality of the Transformer for enhancing referring image segmentation is still far from being sufficiently explored in the conventional paradigm.
Specifically, cross-modal interactions occur only after feature encoding, and a cross-modal decoder is solely responsible for aligning the visual and linguistic features.
As a result, previous methods fail to effectively leverage the rich Transformer layers in the encoder for excavating helpful multi-modal context.
To address these issues, a potential solution is to exploit a visual encoder network for jointly embedding linguistic and visual features during visual encoding.

Accordingly, we propose a \textbf{L}anguage-\textbf{A}ware \textbf{V}ision \textbf{T}ransformer (\textbf{LAVT}) network, in which visual features are encoded together with linguistic features, being ``aware'' of their relevant linguistic context at each spatial location.
As shown in~\cref{fig:1} (b), LAVT makes full use of the multi-stage design in a modern vision Transformer backbone network, leading to a hierarchical language-aware visual encoding scheme.
Specifically, we densely integrate linguistic features into visual features via a pixel-word attention mechanism, which occurs at each stage of the network.
The beneficial vision-language cues are then exploited by the following Transformer blocks, \eg,~\cite{Liu_2021_swin}, in the next encoder stage.
This approach enables us to forgo a complicated cross-modal decoder, since the extracted language-aware visual features can be readily adopted to harvest accurate segmentation masks with a lightweight mask predictor.

To evaluate the effectiveness of the proposed method, we conduct extensive experiments on various mainstream referring image segmentation datasets.
Our LAVT achieves 72.73\%, 62.14\%, 61.24\%, and 60.50\% overall IoU on the validation sets of RefCOCO~\cite{yu2016modeling}, RefCOCO+~\cite{yu2016modeling}, G-Ref (UMD partition)~\cite{nagaraja16refexp}, and G-Ref (Google partition)~\cite{mao2015generation}, improving the state of the art for these datasets by absolute margins of 7.08\%, 6.64\%, 6.84\%, and 8.57\%, respectively.

To summarize, our contributions are twofold: 
\begin{itemize}
\item We propose LAVT, a Transformer-based referring image segmentation framework that performs language-aware visual encoding in place of cross-modal fusion post feature extraction.
\item We achieve new state-of-the-art results on three datasets for referring image segmentation, demonstrating the effectiveness and generality of the proposed method. Source code is available at \href{https://github.com/yz93/LAVT-RIS}{LAVT-RIS}.
\end{itemize}

\section{Related work}
\label{sec:related}

\noindent \textbf{Referring image segmentation}
has attracted growing attention in the research community and there are two main processes in conventional pipelines: (1) extracting features from the text and image inputs respectively, and (2) fusing the multi-modal features to predict the segmentation mask.
In the first process, previous methods adopt recurrent neural networks~\cite{hu2016segmentation, lstm, liu2017recurrentInter, Li2018rrn, Jing_2021_Locate} and language Transformers~\cite{refvos, bert} to encode language inputs.
To encode visual inputs, vanilla fully convolutional networks~\cite{fcn, hu2016segmentation, liu2017recurrentInter}, DeeplabV3~\cite{deeplabv3, Li2018rrn, refvos}, and DarkNet~\cite{darknet, Luo_2020_mcn, Jing_2021_Locate} have been successively employed in previous methods with the purpose of learning discriminative representations.

\begin{figure*}[th]
\centering
\includegraphics[width=0.99\textwidth]{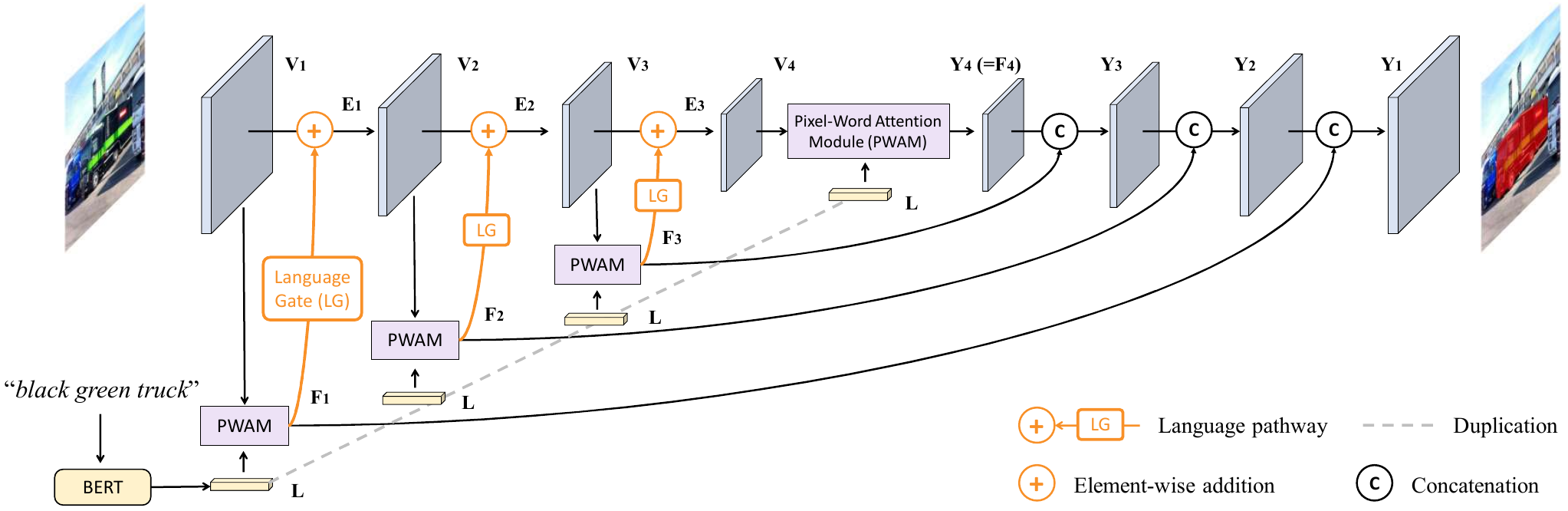}
\vspace{5pt}
\caption{Overall pipeline of the proposed LAVT. We leverage a hierarchical vision Transformer~\cite{Liu_2021_swin} to perform language-aware visual encoding.
At each stage, visual feature maps $V_i$, $i\in \{1,2,3,4\}$ are encoded from the corresponding stage of Transformer layers (which are described in~\cref{sec:lave} and for diagrammatic clarity, are not illustrated in this figure). Then $V_i$ are used as queries for generating a set of position-specific language feature maps $F_i$, $i\in \{1,2,3,4\}$ in the pixel-word attention module (\cref{sec:pwam}). Next, we adaptively fuse $F_i$ with the original $V_i$ via a language pathway (\cref{sec:lp}). The new visual feature maps $E_i$, $i\in \{1,2,3\}$ are then passed into the next stage of Transformer layers for further processing. A standard segmentation decoder head (\cref{sec:segmentation}) produces the final segmentation output.}
\label{fig:2}
\end{figure*}

The multi-modal feature fusion module is the key component that prior arts focus on.
For example, Hu~\etal~\cite{hu2016segmentation} propose the first baseline based on the concatenation operation, which is improved by Liu~\etal~\cite{liu2017recurrentInter} with a recurrent strategy.
Shi~\etal~\cite{shi2018key}, Chen~\etal~\cite{chen2019see}, Ye~\etal~\cite{ye2019cross}, and Hu~\etal~\cite{hu2020brinet} model cross-modal relations between language and vision features via various attention mechanisms.
Yu~\etal~\cite{yu2018mattnet} and Huang~\etal~\cite{huang2020CMPC} leverage knowledge about sentence structures to capture different concepts (\eg, categories, attributes, relations, \etc) in multi-modal features, while Hui~\etal~\cite{hui2020linguistic} exploit syntactic structures among words for guiding multi-modal context aggregation.

The methods most related to ours are VLT~\cite{Ding_2021_vlt} and EFN~\cite{EFN}, where the former designs a Transformer decoder for fusing linguistic and visual features, and the latter adopts a convolutional vision backbone network for encoding language information.
Differently from~\cite{Ding_2021_vlt}, we propose an early fusion scheme which effectively exploits the Transformer encoder for modeling multi-modal context.
Compared to~\cite{EFN}, we do not rely on a complicated cross-modal decoder, leading to a clearer and more effective framework.
Under fair comparisons, our method outperforms these two previous counterparts by large margins.

\noindent \textbf{Transformer}
is first introduced as a sequence-to-sequence deep attention-based language model\cite{attention-all-you-need}, and has dominated the natural language processing (NLP) field~\cite{bert,dai2019transformer,Yang2019xlnet} due to its strong capability on global context modeling.
More recently, it has achieved great success on various computer vision tasks, \eg, image classification~\cite{dosovitskiy2021vit, deit, Liu_2021_swin}, action recognition~\cite{arnab2021vivit, videoswin}, object detection~\cite{detr, Deformable_DETR, Liu_2021_swin}, and semantic segmentation~\cite{Liu_2021_swin, zheng2021rethinking,Segmenter}.

There has also been a rich line of work on Transformers in the intersection area of computer vision and NLP~\cite{rao2021denseclip,kamath2021mdetr}.
For example, Radford~\etal devise a large-scale pretraining model, named CLIP~\cite{clip}, which applies contrastive learning~\cite{hadsell2006dimensionality,sohn2016improved,he2020momentum} on features learned by a vision Transformer and a language Transformer.
Hu~\etal~\cite{hu2021unit} propose a Unified Transformer (UniT) model that jointly learns multiple vision-language tasks across different domains.
Besides, growing efforts have been devoted to other tasks such as visual question answering~\cite{ViLBERT} and text-to-video retrieval~\cite{clipbert}.
However, to the best of our knowledge, there have been very few attempts on designing a unified Transformer model for the task of referring image segmentation.

\section{Method}
\label{sec:method}
\cref{fig:2} illustrates the pipeline of our Language-Aware Vision Transformer (LAVT), which leverages a hierarchical vision Transformer to jointly embed language and vision information to facilitate cross-modal alignments.
In this section, we start by introducing our language-aware visual encoding strategy in~\cref{sec:lave}, which is achieved with a pixel-word attention module detailed in~\cref{sec:pwam} and a language pathway detailed in~\cref{sec:lp}.
Then in~\cref{sec:segmentation} we describe the light-weight mask predictor used to obtain final results.

\subsection{Language-aware visual encoding}
\label{sec:lave}
Given an input pair of an image and a natural language expression that specifies an object from the image, our model outputs a pixel-wise mask that delineates the object.
To extract language features, we employ a deep language representation model to embed the input expression into high-dimensional word vectors.
We denote the language features as $L\in\mathbb{R}^{C_t\times T}$, where $C_t$ and $T$ denote the number of channels and the number of words, respectively.

After obtaining the language features, we perform joint visual feature encoding and vision-language (which is also called ``cross-modal'' or ``multi-modal'' in the following contents) feature fusion through a hierarchy of vision Transformer layers organized into four stages. We index each stage using $i\in \{1,2,3,4\}$ in the bottom-up direction.
Each stage employs a stack of Transformer encoding layers (with the same output size) $\phi_{i}$, a multi-modal feature fusion module $\theta_{i}$, and a learnable gating unit $\psi_{i}$.
Within each stage, language-aware visual features are generated and refined via three steps.
First, the Transformer layers $\phi_{i}$ take the features from the previous stage as input, and output enriched visual features, denoted as $V_i\in \mathbb{R}^{C_{i}\times H_{i}\times W_{i}}$.
Then, $V_i$ are combined with language features $L$ via the multi-modal feature fusion module $\theta_i$ to produce a set of multi-modal features, denoted as $F_i\in \mathbb{R}^{C_{i}\times H_{i}\times W_{i}}$.
Finally, each element in $F_i$ is weighted by the learnable gating unit $\psi_{i}$ and then added element-wise to $V_i$ to produce a set of enhanced visual features embedded with linguistic information, which we denote as $E_i\in \mathbb{R}^{C_{i}\times H_{i}\times W_{i}}$.
We refer to the computations in this final step as the language pathway.
Here, $C_{i}$, $H_{i}$, and $W_{i}$ denote the number of channels, the height, and the width of feature maps in the $i$-th stage, respectively.

The four stages of Transformer encoding layers correspond to the four stages in a Swin Transformer~\cite{Liu_2021_swin}, which is an efficient hierarchical vision backbone suitable for addressing dense prediction tasks.
The multi-modal feature fusion module within each stage is our proposed pixel-word attention module (PWAM), which is designed with the aim to densely align linguistic meanings with visual clues.
And the gating unit is what we refer to as the language gate (LG), a special unit that we devise for regulating the flow of linguistic information along the language pathway (LP).

\begin{figure}[t]
\centering
\includegraphics[width=0.99\columnwidth]{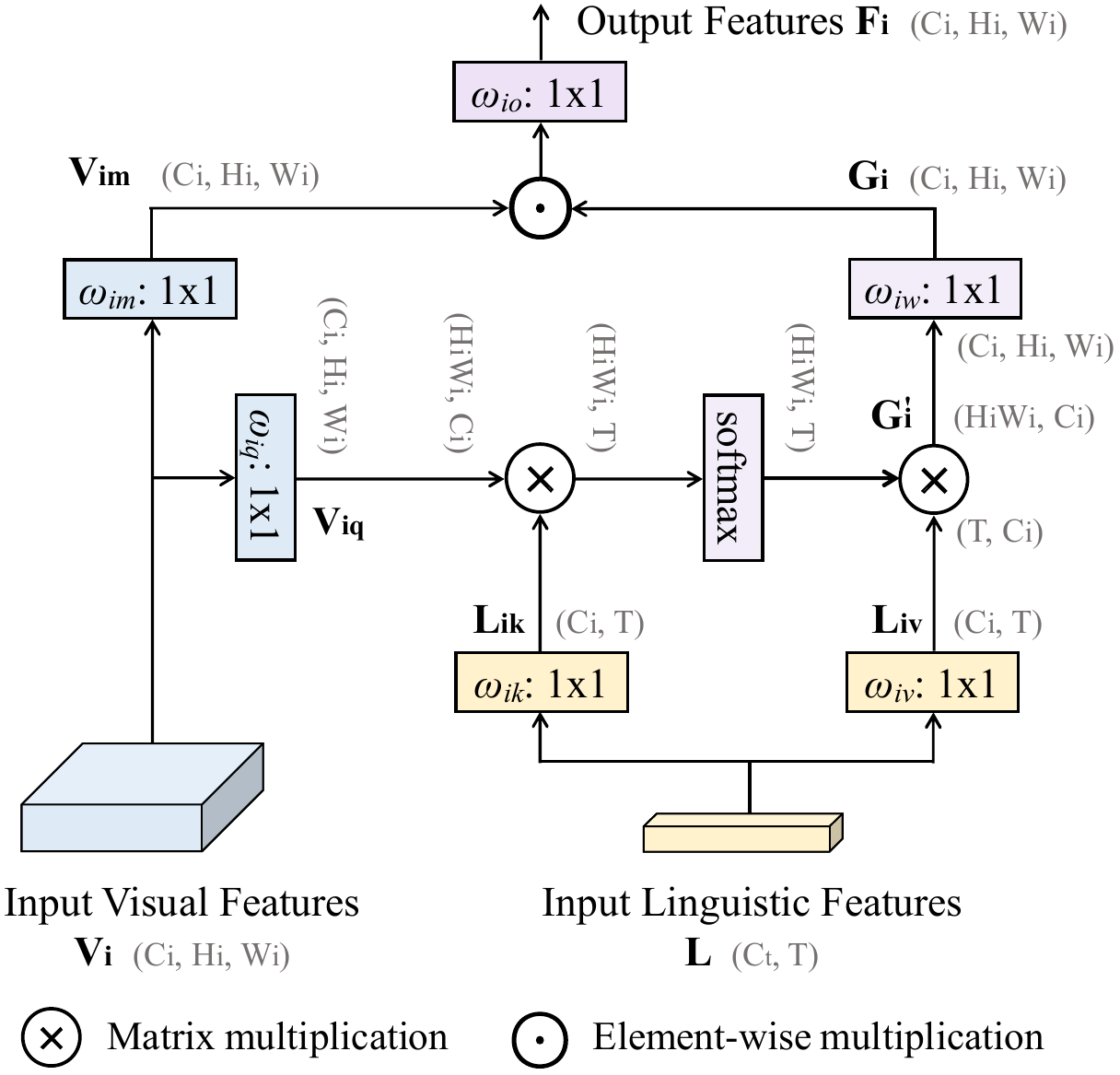}
\vspace{8pt}
\caption{Pipeline of the pixel-word attention module (PWAM). First, a single-head scaled dot-product attention~\cite{attention-all-you-need} is performed using the input visual feature maps $V_i$ as queries and the input linguistic feature maps $L$ as keys and values. The result, $G_i$, is a set of linguistic feature maps of the same spatial size as $V_i$. $G_i$ is then multiplied element-wise with a projection of the input visual feature maps $V_{im}$, followed by another projection before final output. A detail which we found important empirically is the adoption of an instance normalization~\cite{ulyanov2016instance} layer in the projection functions $\omega_{iq}$ and $\omega_{iw}$ (see the text below and~\Cref{tab:3}).}
\label{fig:3}
\end{figure}

\subsection{Pixel-word attention module}
\label{sec:pwam}
In order to separate a target object from its background, it is important to align the visual and linguistic representations of the object across modalities.
One general approach is to combine the representation of each pixel with the representation of the referring expression, and learn multi-modal representations that are discriminative of a ``referent'' class and a ``background'' class.
Previous approaches have developed various mechanisms for addressing this challenge, including dynamic convolutions~\cite{margffoy2018dynamic}, concatenations~\cite{hu2016segmentation,margffoy2018dynamic,Li2018rrn}, cross-modal attentions~\cite{shi2018key,ye2019cross,hu2020brinet,luo2020cascade,EFN}, graph neural networks~\cite{liu2021cmpc}, \etc.
Compared to most of the previous cross-modal attention mechanisms~\cite{shi2018key,ye2019cross,hu2020brinet,luo2020cascade,EFN}, our pixel-word attention module (PWAM) produces a much smaller memory footprint as it avoids computing attention weights between two image-sized spatial feature maps, and is also simpler due to fewer attention steps.

\cref{fig:3} illustrates PWAM schematically.
Given the input visual features $V_i\in \mathbb{R}^{C_i\times H_i\times W_i}$ and linguistic features $L\in \mathbb{R}^{C_t\times T}$, PWAM performs multi-modal fusion in two steps, as introduced in the following.
First, at each spatial location, PWAM aggregates the linguistic features $L$ across the word dimension to generate a position-specific, sentence-level feature vector, which collects linguistic information most relevant to the current local neighborhood.
This step generates a set of spatial feature maps, $G_i\in \mathbb{R}^{C_{i}\times H_{i}\times W_{i}}$.
Concretely, we obtain $G_i$ as follows
\begin{align} \label{eq:1}
V_{iq} &= \mbox{flatten}(\omega_{iq}(V_i)), \\  \label{eq:2}
L_{ik} &= \omega_{ik}(L), \\  \label{eq:3}
L_{iv} &= \omega_{iv}(L), \\  \label{eq:4}
G_i' &= \mbox{softmax}(\frac{V_{iq}^{T}L_{ik}}{\sqrt{C_i}})L_{iv}^{T},\\  \label{eq:5}
G_i &=  \omega_{iw}(\mbox{unflatten}(G_i'^T)),  
\end{align}
where $\omega_{iq}$, $\omega_{ik}$, $\omega_{iv}$, and $\omega_{iw}$ are projection functions.
Each of the language projections $\omega_{ik}$ and $\omega_{iv}$ is implemented as a 1$\times$1 convolution with $C_i$ number of output channels.
And the query projection $\omega_{iq}$ and the final projection $\omega_{iw}$ each is implemented as a 1$\times$1 convolution followed by instance normalization, with $C_i$ number of output channels.
Here, `flatten' refers to the operation of unrolling the two spatial dimensions into one dimension in row-major, C-style order, and `unflatten' refers to the opposite operation.
These two operations and transposing are used to transform feature maps into proper shapes for calculation.
Eqs.~\ref{eq:1}~to~\ref{eq:5} implement the scaled dot-product attention~\cite{attention-all-you-need} using visual features $V_i$ as the query and linguistic features $L$ as the key and the value, with instance normalization after linear transformation in the query projection function $\omega_{iq}$ and the output projection function $\omega_{iw}$.

Second, after obtaining the linguistic features $G_i$ which have the same shape as $V_i$, we combine them to produce a set of multi-modal feature maps $F_i$ via element-wise multiplication.
Specifically, our step is described as follows
\begin{align} \label{eq:6}
V_{im} &= \omega_{im}(V_i), \\  \label{eq:7}
F_i &= \omega_{io}(V_{im} \odot G_i),
\end{align}
where $\odot$ denotes element-wise multiplication and $\omega_{im}$ and $\omega_{io}$ are a visual projection and a final multi-modal projection, respectively.
Each of the two functions is implemented as a 1$\times$1 convolution followed by ReLU~\cite{relu} nonlinearity.

\subsection{Language pathway}
\label{sec:lp}
\begin{figure}[t]
\centering
\includegraphics[width=0.9\columnwidth]{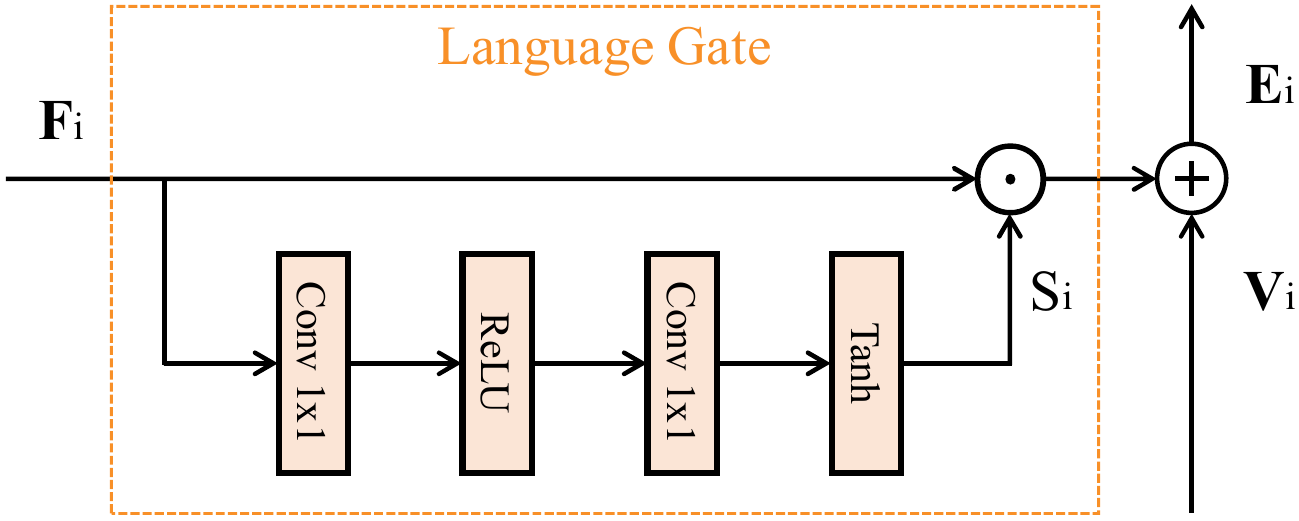}
\vspace{5pt}
\caption{The schema of the language pathway, which leverages a language gate (LG) for controlling multi-modal information flow. LG is implemented as a two-layer perceptron.}
\label{fig:4}
\end{figure}

As described earlier, at each stage, we merge the output from PWAM, $F_i$, with the output from the Transformer layers, $V_i$.
We refer to the computations in this merging operation as the language pathway.
In order to prevent $F_i$ from overwhelming the visual signals in $V_i$ and to allow an adaptive amount of linguistic information flowing to the next stage of Transformer layers, we design a language gate which learns a set of element-wise weight maps based on $F_i$ to re-scale each element in $F_i$.
The language pathway is schematically illustrated in~\cref{fig:4} and mathematically described as follows
\begin{align} \label{eq:8}
S_i &= \gamma_i(F_i), \\  \label{eq:9}
E_i &= S_i \odot F_i + V_i,
\end{align}
where $\odot$ indicates element-wise multiplication and $\gamma_i$ is a two-layer perceptron, with the first layer being a 1$\times$1 convolution followed by ReLU~\cite{relu} nonlinearity and the second layer being a 1$\times$1 convolution followed by a hyperbolic tangent function.
As detailed in the ablation studies in~\Cref{tab:3}, we have experimented with and without using a language gate along the language pathway, as well as different final nonlinear activation functions in the language gate, and found that using the gate with $tanh$ final nonlinearity works the best for our model.
The summation operation in Eq.~\ref{eq:9} is an effective way of utilizing pre-trained vision Transformer layers for multi-modal embedding, as the treatment of multi-modal features as ``supplements'' (or ``residuals'') avoids disrupting the initialization weights pre-trained on pure vision data.
We have observed much worse results in the case of adopting replacement or concatenation.

\subsection{Segmentation}
\label{sec:segmentation}
We combine the multi-modal feature maps, $F_i$, $i\in\{1,2,3,4\}$, in a top-down manner to exploit multi-scale semantics for final segmentation.
The decoding process can be described by the following recursive function
\begin{align} \label{eq:10}
\left\{ 
\begin{array}{ll}
Y_4 &= F_4, \\
Y_{i} &= \rho_{i}([\upsilon(Y_{i+1}); F_{i}]),  \quad i = 3, \; 2, \; 1.
\end{array}
\right.
\end{align}

Here `[ ; ]' denotes feature concatenation along the channel dimension, $\upsilon$ represents upsampling via bilinear interpolation, and $\rho_{i}$ is a projection function implemented as two 3$\times$3 convolutions connected by batch normalization~\cite{ioffe2015batch} and ReLU~\cite{relu} nonlinearity.
The final feature maps, $Y_1$, are projected into two class score maps via a 1$\times$1 convolution.

\subsection{Implementation}
\label{sec:implementation}
We implement our method in PyTorch~\cite{paszke2019pytorch} and use the BERT implementation from HuggingFace's Transformer library~\cite{wolf-2020-transformers}.
The Transformer layers in LAVT are initialized with classification weights pre-trained on ImageNet-22K~\cite{imgnet} from the Swin Transformer~\cite{Liu_2021_swin}.
Our language encoder is the base BERT model with 12 layers and hidden size 768 from~\cite{attention-all-you-need} (hence $C_t$ in~\cref{sec:method} is 768) and is initialized using the official pre-trained weights.
The rest of weights in our model are randomly initialized.
$C_i$ in~\cref{sec:method} is set to 512 and the model is optimized with cross-entropy loss.
Following~\cite{Liu_2021_swin}, we adopt the AdamW~\cite{loshchilov2017adamw} optimizer with weight decay 0.01 and initial learning rate 0.00005 with polynomial learning rate decay.
We train our model for 40 epochs with batch size 32.
We iterate through each object (while randomly sampling one referring expression for it) exactly once in an epoch.
Images are resized to 480$\times$480 and no data augmentation techniques are applied.
During inference, $argmax$ along the channel dimension of the score maps are used as predictions.

\begin{table*}[t]
   \centering
   \setlength{\tabcolsep}{2.5mm}{\begin{tabular}{l|l|c|c|c|c|c|c|c|c|c}
      \toprule[1pt]
      \multirow{2}{*}{Method} &
      Language &
      \multicolumn{3}{c|}{RefCOCO}  & \multicolumn{3}{c|}{RefCOCO+} & \multicolumn{3}{c}{G-Ref} \\
      \cline{3-11}
                                    & Model & val   & test A & test B & val & test A & test B & val (U) & test (U) & val (G)  \\
      \hline
      DMN~\cite{margffoy2018dynamic} & SRU & 49.78 & 54.83 & 45.13 & 38.88 & 44.22 & 32.29 & -     & -     & 36.76 \\
      RRN~\cite{Li2018rrn}          & LSTM & 55.33 & 57.26 & 53.93 & 39.75 & 42.15 & 36.11 & -     & -     & 36.45 \\
      MAttNet~\cite{yu2018mattnet}  & Bi-LSTM & 56.51 & 62.37 & 51.70 & 46.67 & 52.39 & 40.08 & 47.64 & 48.61 & -     \\
      CMSA~\cite{ye2019cross}       & None & 58.32 & 60.61 & 55.09 & 43.76 & 47.60 & 37.89 & -     & -     & 39.98 \\
      CAC~\cite{bmvc2019CAC}        & Bi-LSTM & 58.90 & 61.77 & 53.81 & -     & -     & -       & 46.37 & 46.95 & 44.32 \\
      STEP~\cite{chen2019see}       & Bi-LSTM & 60.04 & 63.46 & 57.97 & 48.19 & 52.33 & 40.41 & -     & -     & 46.40 \\
      BRINet~\cite{hu2020brinet}    & LSTM & 60.98 & 62.99 & 59.21 & 48.17 & 52.32 & 42.11 & -     & -     & 48.04 \\
      CMPC~\cite{huang2020CMPC}     & LSTM & 61.36 & 64.53 & 59.64 & 49.56 & 53.44 & 43.23 & -     & -     & 49.05 \\
      LSCM~\cite{hui2020linguistic} & LSTM & 61.47 & 64.99 & 59.55 & 49.34 & 53.12 & 43.50 & -     & -     & 48.05 \\
      CMPC+~\cite{liu2021cmpc}      & LSTM & 62.47 & 65.08 & 60.82 & 50.25 & 54.04 & 43.47 & -     & -     & 49.89 \\
      MCN~\cite{Luo_2020_mcn}       & Bi-GRU & 62.44 & 64.20 & 59.71 & 50.62 & 54.99 & 44.69 & 49.22 & 49.40 & -     \\
      EFN~\cite{EFN}                & Bi-GRU & 62.76 & 65.69 & 59.67 & 51.50 & 55.24 & 43.01 & - & - & 51.93     \\
      BUSNet~\cite{BUSNet}          & Self-Att & 63.27 & 66.41 & 61.39 & 51.76 & 56.87 & 44.13 & - & - & 50.56 \\
      CGAN~\cite{luo2020cascade}    & Bi-GRU & 64.86 & 68.04 & 62.07 & 51.03 & 55.51 & 44.06 & 51.01 & 51.69 & 46.54 \\
      LTS~\cite{Jing_2021_Locate}   & Bi-GRU & 65.43 & 67.76 & 63.08 & 54.21 & 58.32 & 48.02 & 54.40 & 54.25 & - \\
      VLT~\cite{Ding_2021_vlt}      & Bi-GRU & 65.65 & 68.29 & 62.73 & 55.50 & 59.20 & 49.36 & 52.99 & 56.65 & 49.76 \\
      \hline
      LAVT (Ours) & BERT & \textbf{72.73} & \textbf{75.82} & \textbf{68.79} & \textbf{62.14} & \textbf{68.38} & \textbf{55.10} & \textbf{61.24} & \textbf{62.09} & \textbf{60.50} \\
      \bottomrule[1pt]
   \end{tabular}}
   \caption{Comparison with state-of-the-art methods in terms of overall IoU on three benchmark datasets. U: The UMD partition. G: The Google partition. We refer to the language model of each reference method as the main learnable function that transforms word embeddings before multi-modal feature fusion. Interested readers can refer to the respective papers for embedding initialization and other details.}
   \label{tab:1}
\end{table*}

\section{Experiments}
\label{sec:experiment}
\subsection{Datasets and metrics}
\label{sec:datasets_and_metrics}
We evaluate our method on three standard benchmark datasets, RefCOCO~\cite{yu2016modeling}, RefCOCO+~\cite{yu2016modeling}, and G-Ref~\cite{mao2015generation,nagaraja16refexp}.
Images in the three datasets are collected from the MS~COCO dataset~\cite{lin2014mscoco} and annotated with natural language expressions.
Each of RefCOCO, RefCOCO+, and G-Ref contains \num{19994}, \num{19992}, and \num{26711} images, with \num{50000}, \num{49856}, and \num{54822} annotated objects and \num{142209}, \num{141564}, and \num{104560} annotated expressions, respectively.
Expressions in RefCOCO and RefCOCO+ are very succinct (containing 3.5 words on average).
In contrast, expressions in G-Ref are more complex (containing 8.4 words on average), which makes the dataset particularly challenging.
Conversely, RefCOCO and RefCOCO+ tend to have more objects of the same category per image (3.9 on average) compared to G-Ref (1.6 on average), therefore they better evaluate an algorithm's ability to comprehend instance-level details.
A characteristic of RefCOCO+ is that location words are banned in its expressions, which also makes it more challenging.
Additionally, there are two different partitions of the G-Ref dataset, one by UMD~\cite{nagaraja16refexp} and the other by Google~\cite{mao2015generation}.
We report results on both.
When evaluating on each dataset, we train our model on the training set of that dataset.
Finally, we make note of the ambiguities and foul language found in many expressions of RefCOCO with the hope that future community efforts will address them.

We adopt the common metrics of overall intersection-over-union (oIoU), mean intersection-over-union (mIoU), and precision at the 0.5, 0.7, and 0.9 threshold values.
The overall IoU is measured as the ratio between the total intersection area and the total union area of all test samples, each of which is a language expression and an image.
This metric favors large objects.
The mean IoU is the IoU between the prediction and ground truth averaged across all test samples.
This metric treats large and small objects equally.
The precision metric measures the percentage of test samples that pass an IoU threshold.

\subsection{Comparison with others}
\label{sec:comparison_w_others}
In~\Cref{tab:1}, we evaluate LAVT against the state-of-the-art referring image segmentation methods on the RefCOCO~\cite{yu2016modeling}, RefCOCO+~\cite{yu2016modeling}, and G-Ref~\cite{mao2015generation,nagaraja16refexp} datasets using the oIoU metric.
LAVT outperforms all previous methods on all evaluation subsets of all three datasets.
Compared with the second-best method, VLT~\cite{Ding_2021_vlt}, LAVT achieves higher performance with absolute margins of 7.08\%, 7.53\%, and 6.06\% on the validation, testA, and testB subsets of RefCOCO, respectively.
Similarly, LAVT attains noticeable improvements over the previous state of the art on RefCOCO+ with wide margins of 6.64\%, 9.18\%, and 5.74\% on the validation, testA, and testB subsets, respectively.
On the most challenging G-Ref dataset (which contains significantly longer expressions), LAVT surpasses the respective second-best methods on the validation and test subsets from the UMD partition by absolute margins of 6.84\% and 5.44\%, respectively.
Similarly on the validation set from the Google partition, LAVT outperforms the second-best method EFN~\cite{EFN} by an absolute margin of 8.57\%.
This performance is achieved without using RefCOCO as additional training data in contrast to EFN.

\begin{figure*}[t]
\centering
\includegraphics[width=0.99\textwidth]{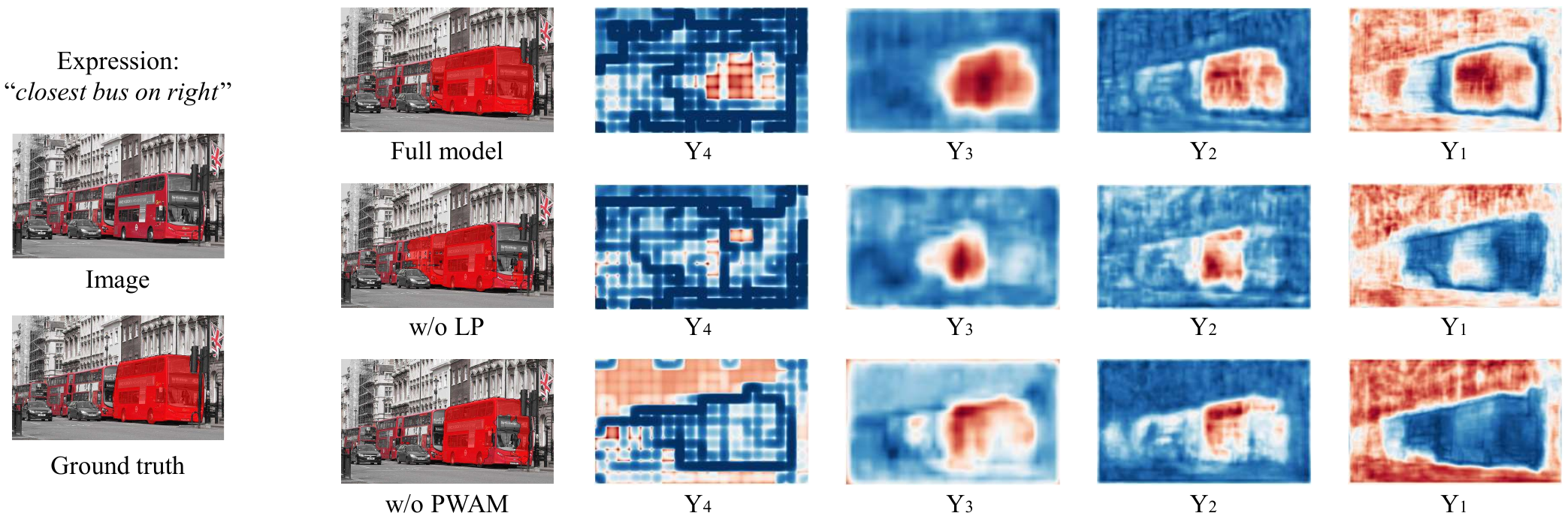}
\caption{Visualized predictions and feature maps on an example from the RefCOCO validation set. From top to bottom, the left-most column illustrates the input expression, the input image, and the ground-truth mask overlaid on the input image. In each row, we visualize the predicted mask and the feature maps used for final classification (\ie, $Y_4$, $Y_3$, $Y_2$, and $Y_1$) from left to right. LP represents the language pathway and PWAM represents the pixel-word attention module.}
\label{fig:5}
\end{figure*}

\subsection{Ablation study}
\label{sec:ablation_study}
We conduct several ablations to evaluate the effectiveness of the key components in our proposed network.

\noindent \textbf{Language pathway (LP).}
\Cref{tab:2} shows that removing LP (which corresponds to, mathematically, the removal of Eqs.~\ref{eq:8} and \ref{eq:9}, or schematically, the removal of the orange stream in~\cref{fig:2}) leads to a drop of 1.95 and 2.50 absolute points in overall IoU and mean IoU, respectively.
In addition, precision drops by 3 to 4 points across all three thresholds.
These results demonstrate the benefit of exploiting our vision Transformer encoder network for jointly embedding linguistic and visual features.

\noindent \textbf{Pixel-word attention module (PWAM).}
In this ablation study, we replace the spatial language feature maps $G_i$ in PWAM with a sentence feature vector globally pooled from all words~\cite{bmvc2021yang}.
As shown in~\Cref{tab:2}, this ablation leads to a drop of 1.70 and 2.15 absolute points in overall IoU and mean IoU, respectively, and a drop of 1 to 2 absolute points in precision across the three thresholds.
These results illustrate the effectiveness of densely aggregating linguistic context via our proposed attention mechanism for enhancing cross-modal alignments.

\begin{table}[t]
   \centering
   \resizebox{\columnwidth}{!}{\begin{tabular}{c|c|c|c|c|c|c}
      \toprule[1pt]
      LP & PWAM & P@0.5 & P@0.7 & P@0.9 & oIoU & mIoU \\
      \hline
      \checkmark & \checkmark & \textbf{84.46} & \textbf{75.28} & \textbf{34.30} & \textbf{72.73} & \textbf{74.46} \\
         & \checkmark & 81.46 & 70.80 & 30.95 & 70.78 & 71.96 \\
      \checkmark &            & 81.76 & 72.76 & 32.46 & 71.03 & 72.31 \\
         &            & 77.87 & 66.93 & 27.95 & 68.82 & 68.87 \\
      \bottomrule[1pt]
   \end{tabular}}
   \caption{Main ablation results on the RefCOCO validation set.}
   \label{tab:2}
\end{table}

\begin{table}[t]
   \centering
   \resizebox{\columnwidth}{!}{\begin{tabular}{l|c|c|c|c|c}
      \toprule[1pt]
      & P@0.5 & P@0.7 & P@0.9 & oIoU & mIoU\\
      \bottomrule[1pt]
      \multicolumn{6}{l}{ \textbf{(a)} activation function in the language gate (LG)} \\
      \toprule[1pt]
      Tanh (*) & \textbf{84.46} & \textbf{75.28} & \textbf{34.30} & \textbf{72.73} & \textbf{74.46} \\
      Sigmoid & 81.89 & 72.71 & 33.35 & 70.49 & 72.47 \\
      \bottomrule[1pt]
      \multicolumn{6}{l}{ \textbf{(b)} normalization layer in pixel-word attention module (PWAM)} \\
      \toprule[1pt]
      InstanceNorm (*) & \textbf{84.46} & \textbf{75.28} & \textbf{34.30} & \textbf{72.73} & \textbf{74.46} \\
      LayerNorm  & 82.97 & 74.15 & 33.99 & 71.92 & 73.32 \\
      BatchNorm & 82.89 & 73.82 & 33.53 & 71.59 & 73.09 \\
      None & 81.91 & 72.73 & 33.11 & 70.66 & 72.34 \\
      \bottomrule[1pt]
      \multicolumn{6}{l}{ \textbf{(c)} features used for final classification} \\
      \toprule[1pt]
      $F_4$, $F_3$, $F_2$, $F_1$ (G*) & \textbf{84.46} & \textbf{75.28} & 34.30 & \textbf{72.73} & \textbf{74.46} \\
      $F_4$, $F_3$, $F_2$, $F_1$ (NG) & 84.00 & 74.96 & 33.47 & 72.24 & 73.94 \\
      \hline
      $E_4$, $E_3$, $E_2$, $E_1$ (G) & 83.84 & 74.96 & 34.48 & 72.06 & 73.98 \\
      $E_4$, $E_3$, $E_2$, $E_1$ (NG) & 84.33 & 74.94 & \textbf{34.77} & 72.27 & 74.12 \\
      \hline
      $V_4$, $V_3$, $V_2$ (G) & 83.36 & 74.47 & 32.61 & 71.38 & 73.29 \\
      $V_4$, $V_3$, $V_2$ (NG) & 83.83 & 74.76 & 32.14 & 72.29 & 73.67 \\
      \bottomrule[1pt]
      \multicolumn{6}{l}{ \textbf{(d)} multi-modal attention module} \\
      \toprule[1pt]
      PWAM (*) & \textbf{84.46} & \textbf{75.28} & \textbf{34.30} & \textbf{72.73} & \textbf{74.46} \\
      BCAM~\cite{hu2020brinet} & 82.26 & 72.81 & 33.31 & 70.19 & 72.42 \\
      GA (GARAN)~\cite{luo2020cascade,Luo_2020_mcn} & 83.22 & 74.09 & 32.71 & 71.20 & 73.16 \\
      \bottomrule[1pt]
   \end{tabular}}
   \caption{Ablation studies on the RefCOCO validation set. (G) indicates that LG is adopted in the language pathway and (NG) indicates the opposite. Rows with (*) indicate default choices.}
   \label{tab:3}
\end{table}

\noindent \textbf{Activation function in the language gate (LG).}
Our proposed LG learns a set of spatial weight maps, which give our network the flexibility to control the flow of language information in the language pathway.
In~\Cref{tab:3} (a), we compare the sigmoid function and the hyperbolic tangent function as the final activation function in LG.
Using the sigmoid function leads to inferior results.

\noindent \textbf{Normalization layer in PWAM.}
As described in~\cref{sec:pwam}, we adopt a final instance normalization layer in the projection functions $\omega_{iq}$ and $\omega_{iw}$ in PWAM.
As we illustrate in~\Cref{tab:3} (b), this particular choice of normalization function has a non-trivial effect.
In addition to instance normalization (our default choice), we experiment with batch normalization, layer normalization, and without having a normalization layer in the functions $\omega_{iq}$ and $\omega_{iw}$.
All three other choices lead to 1 to 2 absolute points drop in the overall IoU and mean IoU metrics.
Among these three choices, using batch normalization or layer normalization produces better results than not using a normalization layer.

\begin{figure*}[th!]
\centering
\includegraphics[width=0.99\textwidth]{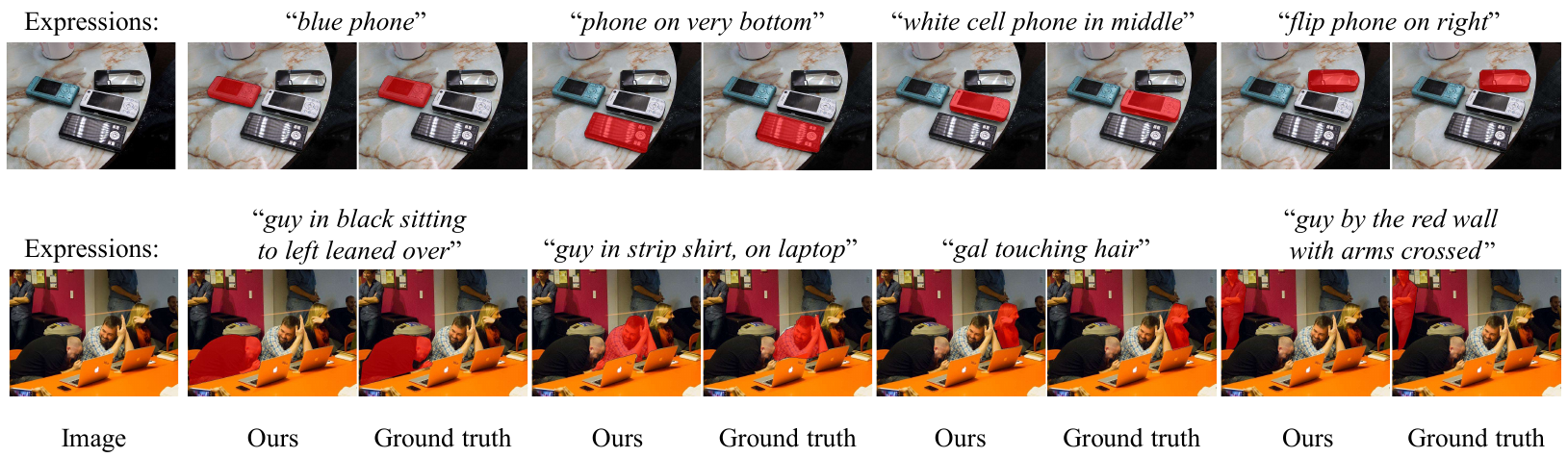}
\caption{Visualizations of our predicted masks and the ground-truth masks on two examples from the RefCOCO validation set.}
\vspace{-0.2cm}
\label{fig:6}
\end{figure*}

\noindent \textbf{Features used for prediction.}
As shown in~\cref{fig:4}, the language-aware visual encoding process of LAVT produces three kinds of spatial feature maps which encapsulate visual and linguistic information, \ie, the outputs from PWAMs ($F_i$, $i \in \{1,2,3,4\}$), the outputs from the Transformer layers ($V_i$, $i \in \{2,3,4\}$), and the inputs to the following Transformer layers ($E_i$, $i \in \{1,2,3\}$).
While our default choice is to use $F_i$ for predicting the object mask, we also consider the other two types of feature maps natural candidates for this purpose.
As shown in~\cref{fig:2}, $E_4$ is not generated in the standard architecture of LAVT.
To have a convincing ablation study, we compute $E_4$ with an additional language pathway as defined in Eqs.~\ref{eq:8} and \ref{eq:9}.
Therefore, we use $E_i$, $i \in \{1,2,3,4\}$ to predict the segmentation masks.
In comparison, as multi-modal information has been progressively integrated into $V_2$, $V_3$, and $V_4$ along the bottom-up computation pathway while $V_1$ contains pure visual information, we do not use $V_1$ for prediction.
In~\Cref{tab:3} (c), we report segmentation results when using each type of features with and without our proposed LG (indicated by ``G'' and ``NG'', respectively).
\Cref{tab:3} (c) shows that using our default choice of $F_i$ with LG produces the best overall results among all choices.
Also, we observe that while LG has a positive effect when using $F_i$ for segmentation, it slightly degrades the results when $E_i$ (72.06\% vs. 72.27\% in oIoU) or $V_i$ (71.38\% vs. 72.29\% in oIoU) are used for segmentation.

\noindent \textbf{Multi-modal attention module.}
In~\Cref{tab:3} (d), we compare PWAM with two state-of-the-art attention modules by directly replacing PWAM with them in our framework, using the same backbone, language model, and training recipes.
Compared to both the grouped attention (GA or GARAN)~\cite{luo2020cascade,Luo_2020_mcn} and the bi-directional cross-modal attention module (BCAM)~\cite{hu2020brinet}, PWAM achieves higher scores across all metrics.
Note that BCAM is representative of the computationally-heavy attention modules and GA is the most recent top-performing module.

\noindent \textbf{Visualized predictions.}
In~\cref{fig:5}, we visualize the predictions and feature maps of our full model and two ablated models (without the language pathway (``w/o LP'') and without the pixel-word attention module (``w/o PWAM''), respectively).
From the first row, we can observe that the higher-level feature maps (\ie, $Y_4$, $Y_3$, $Y_2$) in our full model can accurately locate the semantic concept given in text, while the low-level feature maps (\ie, $Y_1$) contain rich boundary information important to binary segmentation.
Comparing the predicted masks between the three models, we can observe that the removal of LP and the removal of PWAM both lead to false negative predictions on the front window area of the target bus, while the removal of LP additionally results in the false positive identification of the middle bus.
These qualitative results further validate the effectiveness of our proposed LP and PWAM mechanisms.
More example visualizations are shown in~\cref{fig:6}.

\noindent \textbf{Fair comparison with reference methods.}
To further validate the effectiveness of our proposed method of fusing cross-modal information via a vision Transformer encoder network, in~\Cref{tab:4}, we provide fair comparisons between our method and three previous state-of-the-art methods, LTS~\cite{Jing_2021_Locate}, VLT~\cite{Ding_2021_vlt} and EFN~\cite{EFN}.
All models use BERT\textsubscript{BASE} as the language encoder and Swin-B as the vision backbone network, following the same training settings (described in~\cref{sec:implementation}).
While LTS employs a ``locate-then-segment'' pipeline, VLT is representative of methods that employ a cross-modal Transformer decoder.
Conversely, EFN is representative of methods which fuse cross-modal information via an encoder network and additionally rely on a complicated decoder for obtaining the best results.
As shown in~\Cref{tab:4}, our method outperforms LTS, VLT, and EFN on the validation set of RefCOCO across all metrics.
To further verify that our proposed LAVT encoding scheme is more effective than its counterpart cross-modal decoder approach, we combine our approach with VLT by substituting our original light-weight mask predictor with the cross-modal Transformer decoder from VLT.
As shown in this experiment (indicated by ``ours + VLT'' in~\Cref{tab:4}), employing a Transformer decoder to perform additional cross-modal feature fusion after language-aware visual encoding by LAVT generally does not bring extra gains (except a marginal 0.11\% improvement in P@0.5).

\begin{table}[t]
   \centering
   \resizebox{\columnwidth}{!}{\begin{tabular}{l|c|c|c|c|c}
      \toprule[1pt]
      Method & P@0.5 & P@0.7 & P@0.9 & oIoU & mIoU\\
      \hline
      LTS (Swin-B+BERT)~\cite{Jing_2021_Locate}  & 80.59 & 69.48 & 26.13 & 69.94 & 70.56 \\
      EFN (Swin-B+BERT)~\cite{EFN}  & 82.55 & 73.27 & 31.68 & 70.76 & 72.95 \\
      VLT (Swin-B+BERT)~\cite{Ding_2021_vlt}  & 83.24 & 72.81 & 24.64 & 70.89 & 71.98\\
      Ours + VLT~\cite{Ding_2021_vlt} & \textbf{84.57} & 75.14 & 26.36 & 72.12 & 73.57 \\
      \hline
      Ours  & 84.46 & \textbf{75.28} & \textbf{34.30} & \textbf{72.73} & \textbf{74.46}\\
      \bottomrule[1pt]
   \end{tabular}}
   \caption{Comparison between our method, LTS~\cite{Jing_2021_Locate}, VLT~\cite{Ding_2021_vlt}, and EFN~\cite{EFN} on the RefCOCO validation set, where all models use the same backbone, language model, and training recipes.}
   \vspace{-0.2cm}
   \label{tab:4}
\end{table}

\section{Conclusion}
\label{sec:conclusion}
In this paper, we have proposed a Language-Aware Vision Transformer (LAVT) framework for referring image segmentation, which leverages the multi-stage design of a vision Transformer for jointly encoding multi-modal inputs.
Experimental results on three benchmarks have demonstrated its advantage with respect to the state of the art.

\vspace{1ex}\noindent\textbf{Acknowledgements}.
This work is supported by the UKRI grant: Turing AI Fellowship EP/W002981/1, EPSRC/MURI grant: EP/N019474/1, Shanghai Committee of Science and Technology, China (Grant No. 20DZ1100800), and HKU Startup Fund.
We would also like to thank the Royal Academy of Engineering, Tencent, and FiveAI.

\clearpage
{\small
\bibliographystyle{ieee_fullname}
\bibliography{main}
}
\appendix

\section{Potential biases of the language model} \label{sec:lm}
We note that the pre-trained language model BERT~\cite{bert} (which we employ) has been reported as containing ethnic biases of potential societal concern in some studies.
We refer interested readers to the recent work of Ahn~\etal~\cite{ahn2021mitigating} for more details, in which different kinds (including racial, gender, geological, \etc) of ethnic biases are analyzed and mitigation methods are proposed.

\section{The language pathway} \label{sec:lpa}
For the design of our language pathway, we wanted to find a way to allow the vision Transformer layers to embed multi-modal information effectively.
As a result, we built the language pathway as a residual connection~\cite{fpn,he2016deep}, which has been shown effective for combining features containing different types of information in a deep neural network.
And the design of the language gate is inspired by previous work that featured learnable gates for regulating information flow in deep neural networks, such as the LSTM~\cite{lstm}, the SENet~\cite{hu2018squeeze}, and CFBI~\cite{yang2020CFBI}.

\begin{table}[h]
   \centering
   \resizebox{\columnwidth}{!}{
   \begin{tabular}{l|c|c|c|c|c}
      \toprule[1pt]
      Method & P@0.5 & P@0.7 & P@0.9 & oIoU & mIoU\\
      \hline
      *Replacement (w/o LG)  & -- & -- & -- & -- & -- \\
      *Concatenation (w/o LG)  & 72.89 & 58.15 & 20.02 & 60.52 & 63.41 \\
      Sum (w/o LG)  & 84.00 & 74.96 & 33.47 & 72.24 & 73.94 \\
      \hline
      Sum (with LG; default)  & \textbf{84.46} & \textbf{75.28} & \textbf{34.30} & \textbf{72.73} & \textbf{74.46}\\
      \bottomrule[1pt]
   \end{tabular}
   }
   \caption{Design alternatives for the language pathway (annotated with the asterisk). `LG' is short for language gate. `--' indicates that training suffered extremely slow convergence.}
   \label{tab:5}
\end{table}

\section{Precision-recall analysis} \label{sec:pr}
\begin{figure}[t]
\centering
\includegraphics[width=0.99\columnwidth]{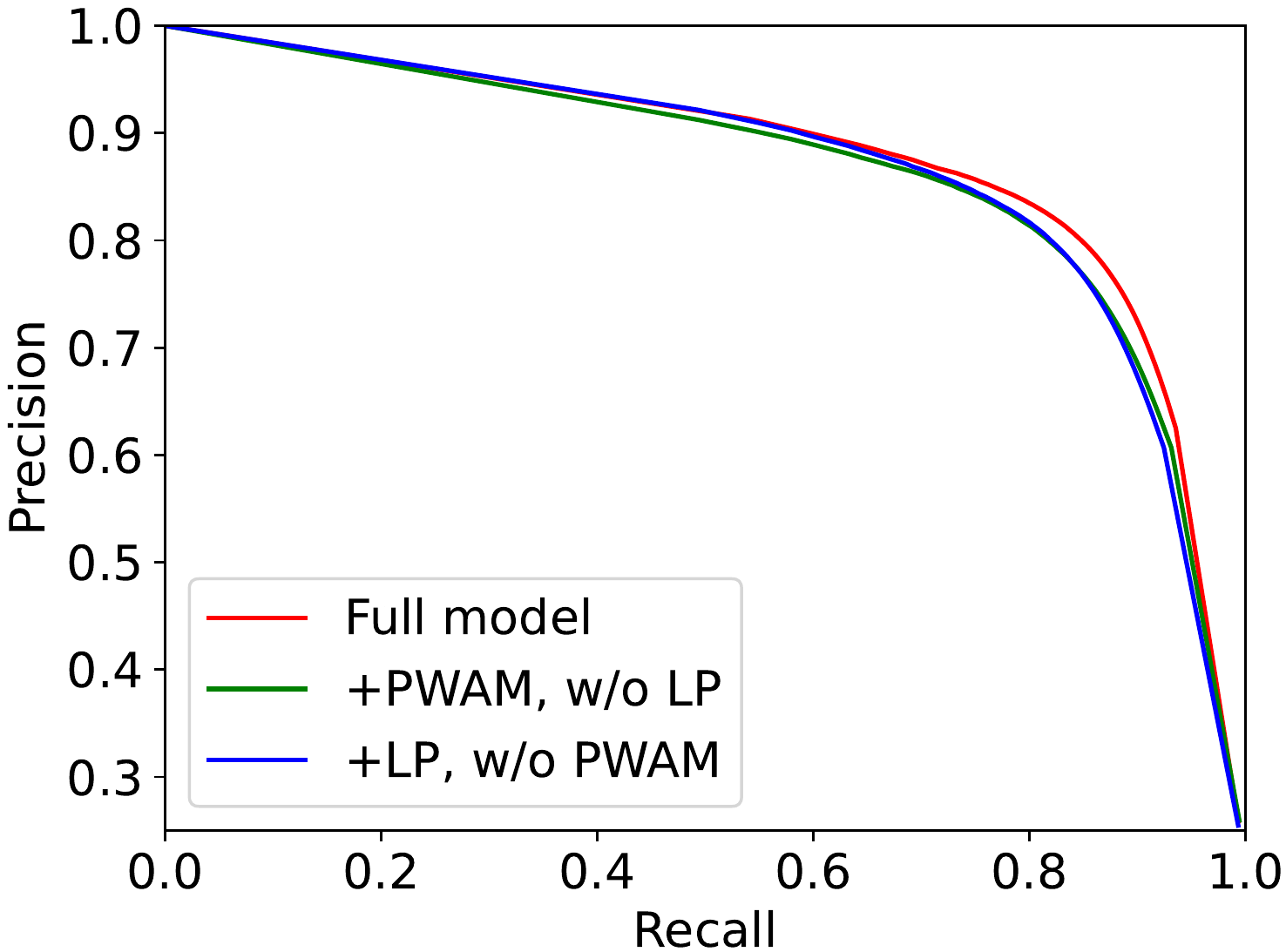}
\caption{Precision-recall (PR) curves on the RefCOCO validation set. The full model obtains the best PR trade-off compared to the ablated models.
Between the ``+LP'' model (blue) and the ``+PWAM'' model (green), a close observation will show that LP maintains a slight advantage in precision over PWAM up until around 0.8 recall.}
\label{fig:7}
\end{figure}

To understand the precision-recall trade-off of LAVT and two of its ablated models, in~\cref{fig:7} we compute and plot the average precision and the average recall of all test samples in the validation set of RefCOCO at 100 thresholds evenly spaced out from 0 to 1 (where the prediction for a pixel is positive if the softmax-normalized score map of the object class exceeds the threshold and is negative otherwise).

\newpage
\onecolumn
\section{Mean IoU} \label{sec:mIoU}
\begin{table*}[ht]
   \centering
   \setlength{\tabcolsep}{2.5mm}{\begin{tabular}{l|l|c|c|c|c|c|c|c|c|c}
      \toprule[1pt]
      \multirow{2}{*}{Method} &
      Language &
      \multicolumn{3}{c|}{RefCOCO}  & \multicolumn{3}{c|}{RefCOCO+} & \multicolumn{3}{c}{G-Ref} \\
      \cline{3-11}
                                    & Model & val   & test A & test B & val & test A & test B & val (U) & test (U) & val (G)  \\
      \hline
      LAVT (Ours) & BERT & 74.46 & 76.89 & 70.94 & 65.81 & 70.97 & 59.23 & 63.34 & 63.62 & 63.66 \\
      \bottomrule[1pt]
   \end{tabular}}
   \caption{Mean IoU of LAVT on the three benchmark datasets. These results complement the overall IoU reported in Table 1 of the main paper. Since mean IoU treats each object equally and does not favor large objects (as overall IoU does), we consider it a fairer metric and recommend more of its use for evaluating this task in the future.}
   \label{tab:6}
\end{table*}

\section{Visualizations} \label{sec:vis}
\begin{figure*}[ht]
\centering
\includegraphics[width=0.99\textwidth]{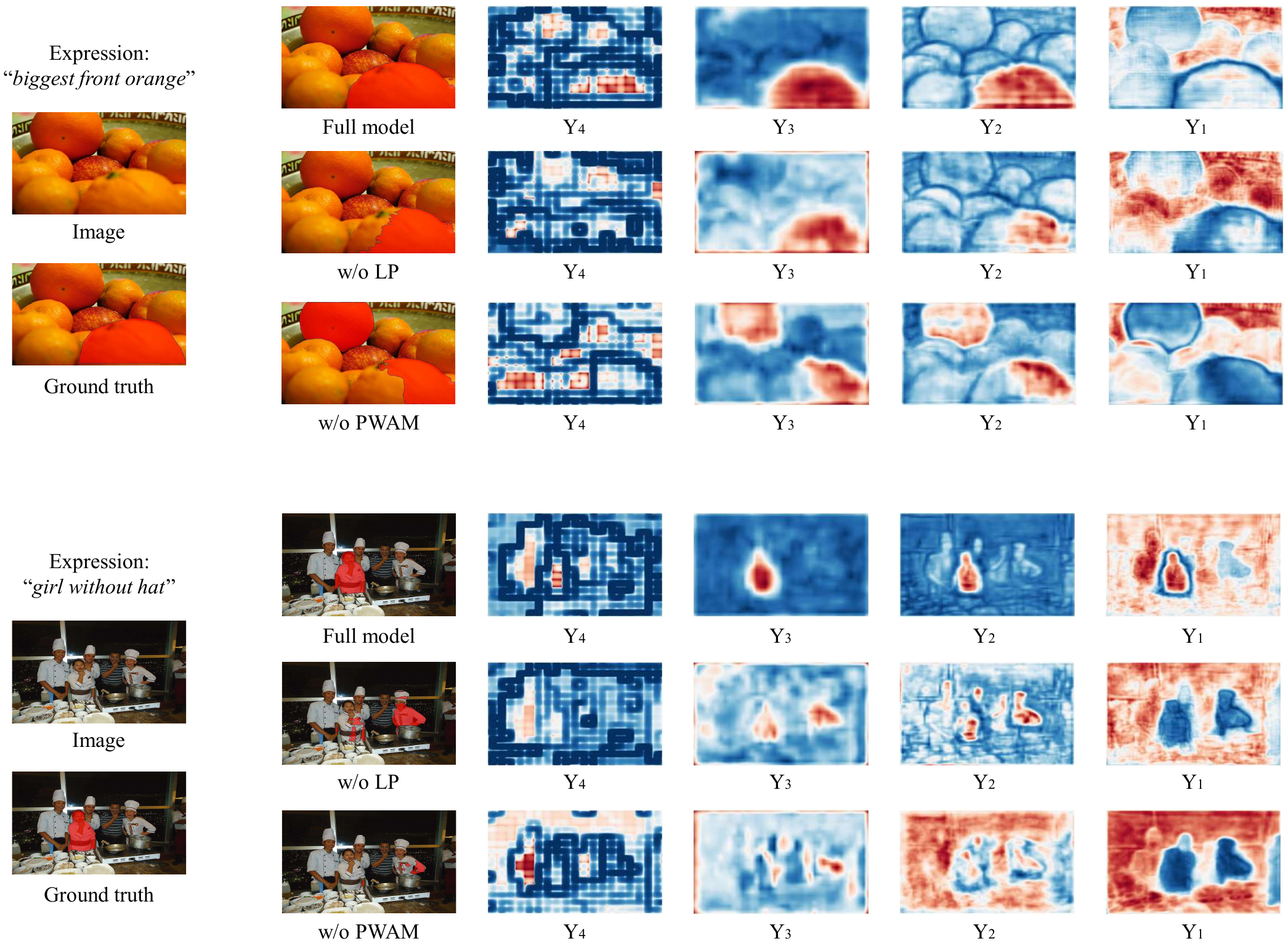}
\caption{Additional visualizations of predictions and feature maps from the RefCOCO validation set. For each example, the left-most column illustrates the input expression, the input image, and the ground-truth mask overlaid on the input image. In each row, we visualize the predicted mask and the feature maps used for final classification (\ie, $Y_4$, $Y_3$, $Y_2$, and $Y_1$) from left to right. LP represents the language pathway and PWAM represents the pixel-word attention module.}
\label{fig:8}
\end{figure*}

\begin{figure*}[ht]
\centering
\includegraphics[width=0.99\textwidth]{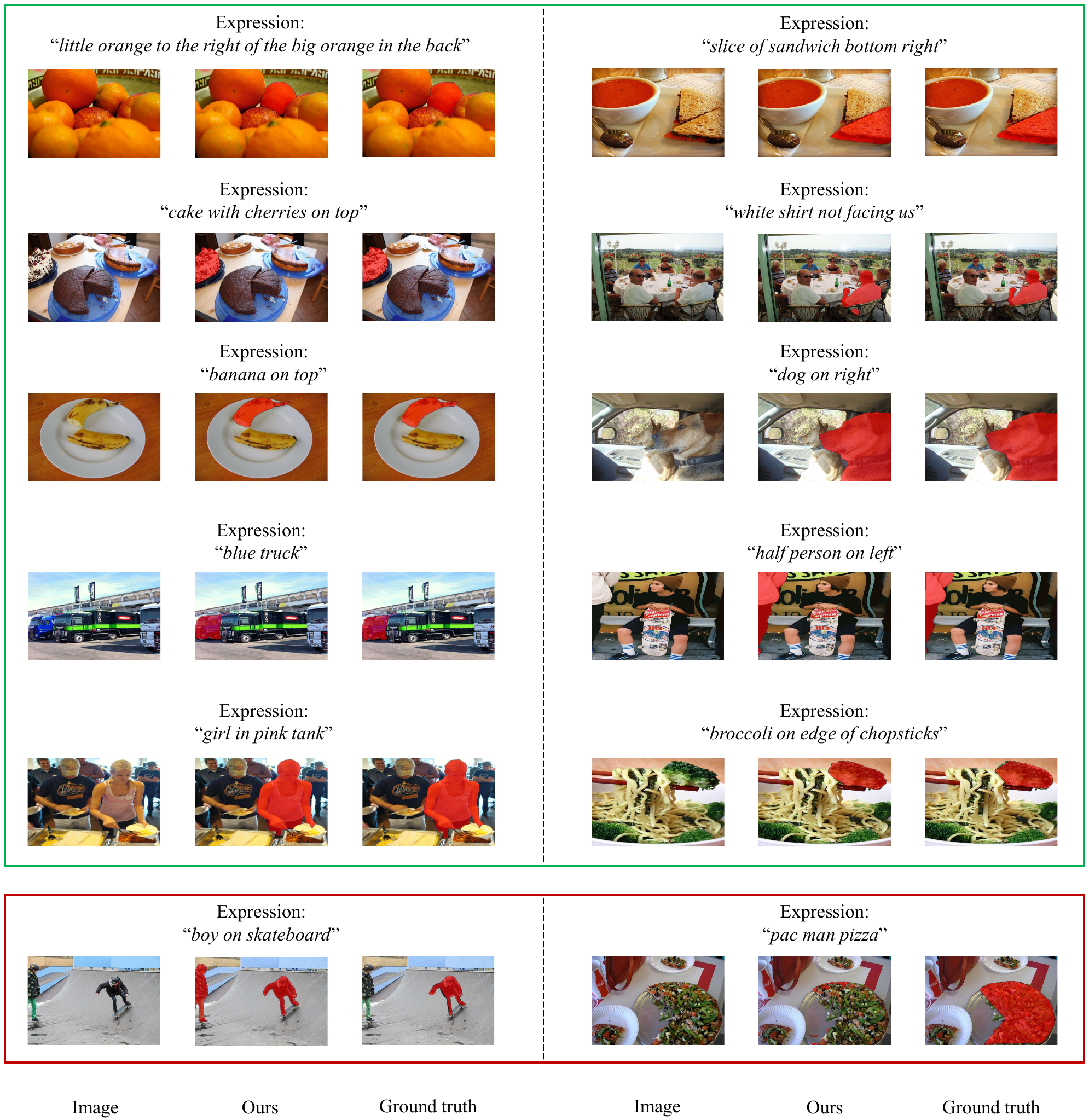}
\caption{Visualizations of our predicted masks and the ground-truth masks on examples from the RefCOCO validation set. Examples enclosed with green lines are successful cases, and those enclosed with red lines are failed cases. In the successful cases, our predictions are nearly identical to the ground truth and are sometimes more accurate than the ground truth (see the second example from the right column, where part of the body of the man behind the chair is missing in the annotation). Among the two demonstrated failure cases, the first one is caused by ambiguity in the given expression (there are two boys that are on a skateboard and our model segments out both) and the second one is caused by our model's lack of knowledge of what a ``pac man'' is (obviously having not played the game Pac-Man, our model fails to associate the shape of the pizza to the shape of a Pac-Man).}
\label{fig:9}
\end{figure*}

\end{document}